\documentclass[letterpaper, 10pt, conference]{ieeeconf}
\IEEEoverridecommandlockouts
\usepackage{multicol}
\usepackage{multirow}

\usepackage{epsfig}
\usepackage{mathptmx}
\usepackage{times}
\usepackage{float}
\usepackage{color}
\usepackage{balance}      

\usepackage{psfrag}
\usepackage{url}
\usepackage{stfloats}
\usepackage{bbold}

\usepackage[utf8]{inputenc}
\usepackage{amssymb}
\usepackage{mathrsfs}
\usepackage{amsmath}
\usepackage{accents}

\usepackage{amsthm}

\usepackage{cuted}           
\let\labelindent\relax
\usepackage{enumitem}        
\setenumerate[enumerate]{align=left}

\usepackage{algorithm}       
\usepackage{algpseudocode}   
\usepackage{ifsym}
\usepackage{mathtools}

\usepackage{cite}
\usepackage{hyperref}

\theoremstyle{plain}

\theoremstyle{definition}

\theoremstyle{definition}

\theoremstyle{plain}

\newcommand{\norm}[1]{\left\lVert#1\right\rVert}
\newcommand{\abs}[1]{\left\lvert#1\right\rvert}

\usepackage{xpatch}
\makeatletter
\xpatchcmd{\@thm}{\thm@headpunct{.}}{\thm@headpunct{}}{}{}
\makeatother
\xpatchcmd{\thm}{\@addpunct{.}}{\@addpunct{:}}{}{}

\makeatletter
\xpatchcmd{\@lemma}{\lemma@headpunct{.}}{\lemma@headpunct{}}{}{}
\makeatother

\hypersetup{
    colorlinks=true,
    linkcolor=blue,
    filecolor=blue,      
    urlcolor=blue,
    citecolor=blue,
}

\title{\bf 
Feasible Computationally Efficient Path Planning for UAV Collision Avoidance
}

\author{Han Wang, Muqing Cao, Hao Jiang and Lihua Xie
\thanks{
The work is partially supported by ST Engineering - NTU Corporate Lab under the NRF Corporate Lab @ University Scheme. 
}
\thanks{
H. Wang, M. Cao and L. Xie are with the School of Electrical and Electronic Engineering, Nanyang Technological University, 50 Nanyang Avenue, Singapore (e-mail:\{wang.han,mqcao,elhxie\}@ntu.edu.sg). 
}
\thanks{
H. Jiang is with College of Electrical Engineering and Automation, Fuzhou University, Fuzhou, China (e-mail:jiangh@fzu.edu.cn) and was with the School of Electrical and Electronic Engineering, Nanyang Technological University, 50 Nanyang Avenue, Singapore.
}
}
\begin{document}

\maketitle

\begin{abstract}
This paper presents a robust computationally efficient real-time collision avoidance algorithm for Unmanned Aerial Vehicle (UAV), namely Memory-based Wall Following-Artificial Potential Field (MWF-APF) method. The new algorithm switches between Wall-Following Method (WFM) and Artificial Potential Field method (APF) with improved situation awareness capability. Historical trajectory is taken into account to avoid repetitive wrong decision. Furthermore, it can be effectively applied to platform with low computing capability. As an example, a quad-rotor equipped with limited number of Time-of-Flight (TOF) rangefinders is adopted to validate the effectiveness and efficiency of this algorithm. Both software simulation and physical flight test have been conducted to demonstrate the capability of the MWF-APF method in complex scenarios.
\end{abstract}

\section{Introduction} \label{sec:intro}
\begin{figure}[!h]
\centering
    \includegraphics[width=0.75\linewidth]{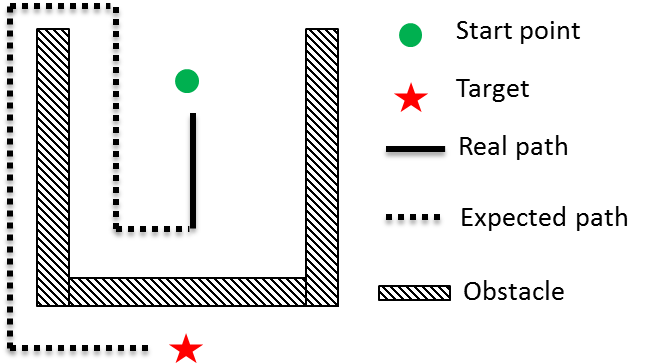}
    \caption{Local Minimum of APF}
    \label{fig:localmin}
\end{figure}

Unmanned Aerial Vehicle (UAV) is one of the most promising technologies in the 21st century. Thanks to its low cost and high mobility, UAV has the potential to replace human in jobs such as package delivery, forest rescue and so forth. Moreover, heavy lift drones have been employed as an innovative alternative of transportation. However, with the popularization of UAV, there are increasing reports of accidental vehicle crash of UAV due to poor pilot control. Safety concern is real and it becomes the bottleneck of UAV industry. Collision avoidance is the key for a successful navigation mission. In recent years, researchers have been working on collision detection and avoidance system (CDAS) that can be implemented on UAVs \cite{gageik2015obstacle,park2012stereo,kwag2007uav}. A good collision avoidance system is the key to further development on UAV such as swarm system\cite{zhu2017survey}. 
\\
A comprehensive collision avoidance solution includes two parts, namely obstacle detection and obstacle avoidance. Obstacle detection studies different approaches to detect the surrounding obstacles. Accuracy of the generated obstacle profile is the most crucial factor of obstacle detection. It has significant impact on the later part of obstacle avoidance. However, there is no uniform obstacle detection solution at the current stage. In the past decades, researchers are putting effort on Obstacle Avoidance system with various sensing units such as infrared sensor\cite{gageik2015obstacle}, vision sensor\cite{park2012stereo} and radar sensor\cite{kwag2007uav}. Every method has its own advantages and disadvantages \cite{pham2015survey}. Vision-based obstacle detection system is of great interest to research community due to the richness of environment information that can be extracted from digital images. However, the performance of such system is limited by the processing power of the on-board computer equipped by UAV. 
In contrast, a rangefinder such as infrared sensor, does not require high processing power and is able to provide very accurate feedback in terms of distance. The constraint of a rangefinder is the narrow field of view (FOV). Each rangefinder is able to cover 1 to 2 degree FOV only. As for radar-based system, it has much wider FOV. However, there is a higher chance for false detection to occur as the outline of obstacle cannot be well perceived. 
\\
Obstacle avoidance is the task of navigating robot subject to non-collision constraints. Different motion planning methods are used to achieve the task.  
One of the most classic motion planning methods is called artificial potential field (APF) method\cite{andrews1983impedance}, which uses the idea of imaginary force. Previous study\cite{koren1991potential} has shown that APF can provide a computationally efficient motion planning for robot. However, one problem generated by APF is the Local Minimum Problem illustrated in Fig. \ref{fig:localmin}. The robot may be trapped at local minimum points where the imaginary potential force equals to zero and unable to reach the target \cite{koren1991potential}.

\begin{figure}[!h]
\centering
    \includegraphics[width=0.80\linewidth]{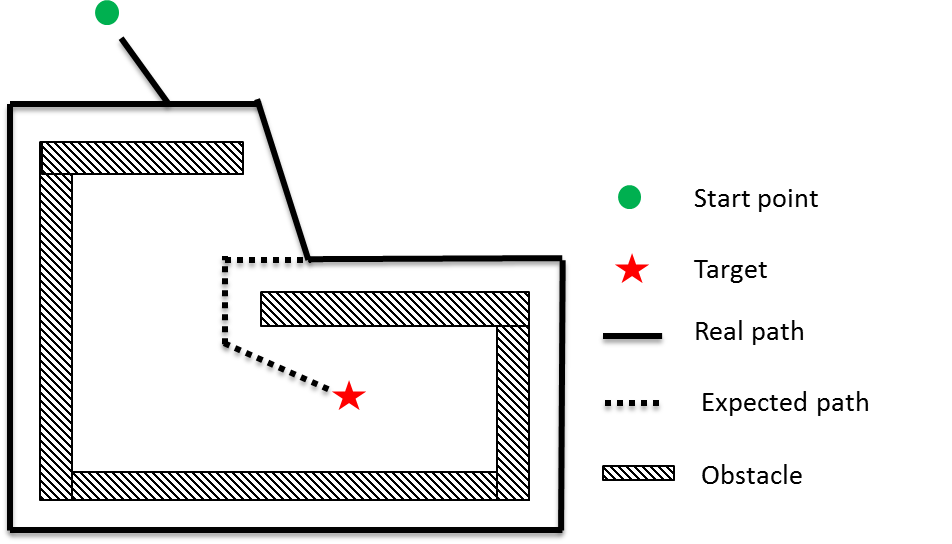}
    \caption{Endless Loop Scenario for Wall-Following Method.}
    \label{fig:endlessloop}
\end{figure}

Another popular motion planning method is called Bug Algorithm. It was firstly introduced by Lumelsky and Stepanov\cite{lumelsky1986dynamic}. This method adopts the strategy to follow the path specified by the contour of an obstacle. Similar to APF, it is an effective and efficient solution for the obstacle avoidance. Research on the Bug Algorithm has made rapid progress in recent years and evolved into methods such as I-Bug Algorithm \cite{taylor2009bug} , Point Bug Algorithm\cite{buniyamin2011simple}, Insert Bug Algorithm\cite{xu2013vectorization}, Vis Bug Algorithm\cite{lumelsky1990incorporating} and Dist Bug Algorithm \cite{kamon1997sensory}. All these algorithms are categorized as Bug Family Algorithm (BFA). The Bug Family Algorithm usually has following three assumptions: i) the robot is a point, ii) it has a perfect localization, and iii) its sensors are precise. Due to these assumptions, research on Bug Algorithm mainly focuses on theoretical studies rather than real life implementation. In the real application, we use the wall-following concept more often than bug family algorithm \cite{finkelstein2008wall,gavrilut2008wall}. Wall-Following Method (WFM) is developed based on the key idea of the bug family algorithm with the consideration of real life constraints such as robot size. WFM only requires to estimate the normal and tangent of the obstacle surface from real-time sensor input. UAVs with WFM always keep a safety normal distance from the obstacle and navigate in the tangent direction only. Feasible implementation makes WFM popular in robotics application. However, it is usually used in simple environment. One significant disadvantage of WFM is that it can lead to endless loop or repetitive trajectory in some complex cases shown in Fig. \ref{fig:endlessloop}. If the robot always follows the right direction of the wall tangent when there are obstacles detected, it can never reach the target. 
It is described above that both methods have specific failure senario. However, it is unlikely for both methods to fail at the same time. Hence, researchers have been looking for a method combining both WFM and APF features. These two methods will compensate each other once equipped with appropriate situation awareness. The failure of one algorithm will trigger another algorithm to take over path planning task. The research on this combined method has proven that it can result in better path than individual method \cite{wang2013hybrid, zhu2009improved, yun1997wall}.

\section{Related works that involves combined methods}
As mentioned above, combined WFM and APF method aims to allow each stand-alone method to compensate another. The key point of this method is the situation awareness to make a right choice between two stand-alone methods. A proper situation awareness is able to recognize accurately when either WFM or APF has reached its failure mode. As a result, a better performance can be achieved. 

\subsection{Early stage of situation awareness}
\begin{figure}[!h]
\centering
    \includegraphics[width=0.80\linewidth]{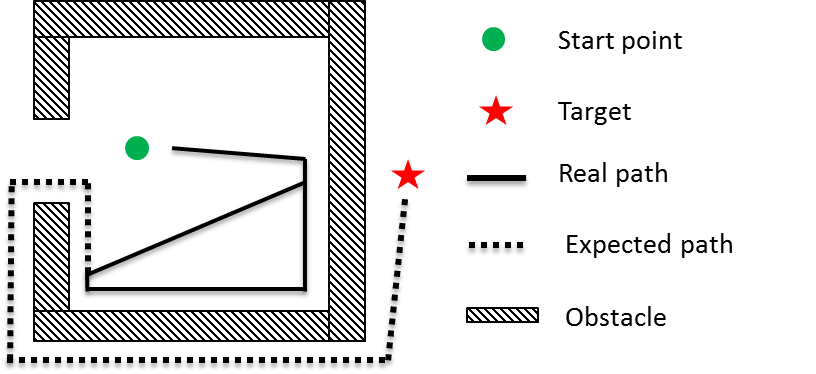}
    \caption{Local Trap Issue for VFF-WFM method.}
    \label{fig:memorylesswfmtrap}
\end{figure}

The idea of combining WFM and APF was initially introduced by\cite{yun1997wall}. This work applied a simple situation awareness for robot to switch between two modes, namely VFF (Virtual Force Field, an extension from APF) and WFM. To switch from VFF to WFM, the following logic was used to distinguish local minimum condition:
\begin{equation}
    \norm{F_{total}}\leq F_{th},
\end{equation}
where $F_{total}$ is the total artificial potential force calculated at the current robot position, $F_{th}$ is the tolerance for a point to be considered as a local minimum point. The condition checks if the robot reaches the local minimum point, that is, the failure of APF happens. If the condition occurs, the system will switch to WFM to escape from the local minimum problem. To switch to VFF from WFM the following condition was used:
\begin{equation}
    \abs{\theta_{goal}-\theta_0}>\frac{\pi}{2},
\end{equation}
where $\theta_{goal}$ and $\theta_0$ are the robot-to-target angle and current tangent of wall respectively. This condition triggers the mode to APF if the difference between robot-to-goal angle and wall-following angle is greater than $\frac{\pi}{2}$. In other words, it happens when robot is moving away from the target. 
\\
This work is useful in an environment where the obstacle profile is simple. However, the failure case happens for complex environment as shown in Fig. \ref{fig:memorylesswfmtrap}. A closed room with small exit opposite to the goal direction is given and the robot is supposed to escape from the room and to move to the goal. The dashed and solid line represents the expected path and resulted path respectively. In the end, the robot will keep endlessly circling along the solid trajectory and fail to navigate to the target point. The root cause of this type of failure is the lack of self-awareness of the looping path. 
\subsection{Modified situation awareness with historical trajectory look-back}
\begin{figure}[!h]
\centering
    \includegraphics[width=0.80\linewidth]{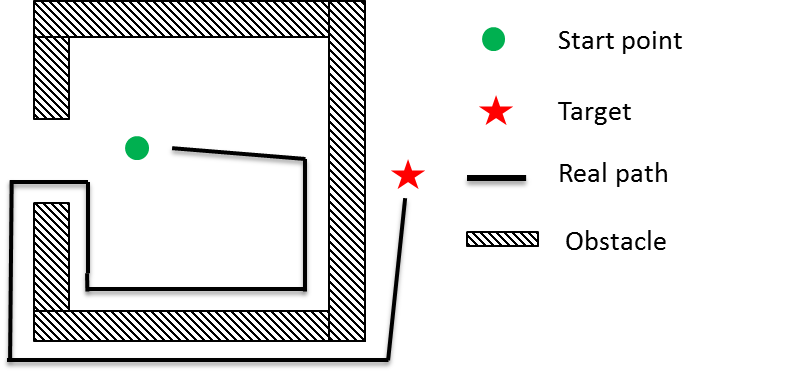}
    \caption{Improved Path Planning with History Look-back.}
    \label{fig:improvedwfm1}
\end{figure}

\begin{figure}[!h]
\centering
    \includegraphics[width=0.80\linewidth]{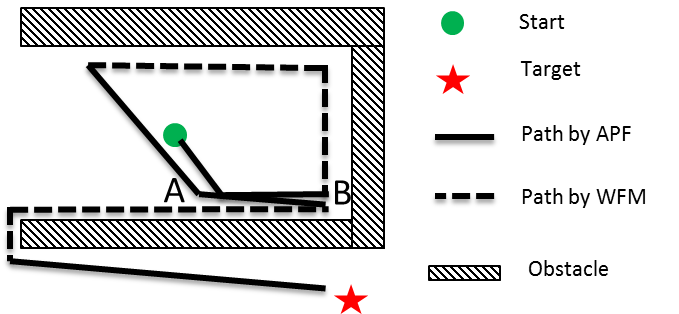}
    \caption{Repetitive Path Generated by Improved Path Planning.}
    \label{fig:improvedwfm}
\end{figure}

\cite{zhu2009improved} reviewed the limitation of \cite{yun1997wall} and improved situation awareness with the look-back of historical trajectory. This paper argued that, without learning from the historical experience, the robot can make the same wrong decision repetitively. Hence in this work, the proposed situation awareness takes the visited locations into consideration. A new condition is added when switching from APF to WFM with comparison between the robot-to-target path and historical trajectory. The robot-to-target path is the straight line path from current position to the goal. The existence of intersection of the robot-to-target path and historical trajectory will prevent robot from switching to APF mode. Fig. \ref{fig:improvedwfm1} shows that the method can resolve the endless loop problem described in \cite{yun1997wall} with the same environment setup. 
\\
However, this work does not make full use of the historical information, which can result in less efficient trajectory. An example is given in Fig. \ref{fig:improvedwfm} where paths navigated by APF and WFM are represented by solid and dashed line respectively. When robot revisits this local minimum point, direction of wall following is changed so that the robot can move to the target point finally. However, repetitive path happens between point A and B marked in Fig. \ref{fig:improvedwfm}. This is because the above-mentioned situation awareness only checks the historical trajectory during WFM mode but not in APF mode. Hence, in the APF mode the robot does not perform checks for repeating decision. We believe that with a better decision mechanism, the trajectory can be further optimized to avoid the path between point A and B as shown in Fig. \ref{fig:improvedwfm}. 

\section{Proposed MWF-APF method}
As discussed above, the look-back of historical trajectory plays an important role in the success of situation awareness system. Based on this, a new method is developed in this paper with memory based Wall-Following Method and Artificial Potential Field method, namely, MWF-APF method. More historical information will be considered in this enhanced situation awareness system.

\subsection{Historical Trajectory Information}
Similar to \cite{zhu2009improved}, the historical trajectory is stored as a series of key frames. Each key frame consists of time $t$, location information $p(t)$  as well as current moving direction $v(t)$. In addition, a flag highlighting those key frames with local minimum of APF is added as extra information for the ease of processing. With storage space constraints in practical application, a new key frame will only be recorded when it satisfies any one of the three conditions below:
\begin{enumerate} [label=\roman*.]
    \item $ \norm{p - p_{his}}>d_{th}$,
    \item $  \norm{p - p_{his}}<d_{th}  $ but $\abs{ \angle(v,v_{his}) }>\theta_{th}  $,
    \item $\norm{F_{total}}\leq F_{th}$,
\end{enumerate}
where $ p_{his} $, $v_{his}$ are the historical position and moving direction stored in the database and $d_{th}$, $\theta_{th}$ are the constants threshold to determine the data size. Constant $d_{th}$ determines how frequently the key frame is updated. A frame is recorded when the distances between current robot position and all other key frame locations are greater than this threshold. If current position coincides with historical position but with moving direction difference greater than $\theta_{th}$, the current status will also be considered as a key frame. So when the robot revisits historical position but with a different moving direction, current status will be recorded as well. In addition, when local minimum point is found, the current status will be automatically considered as a key frame with the highlight of local minimum.

\subsection{Artificial Potential Field}
In APF mode, a virtual force is generated by a combination of both attractive and repulsive force. Assuming that the target position of the robot is $P_t$ while the current position of the robot is $P$, then the attractive potential function takes a similar form as in \cite{yun1997wall}:
\begin{equation} \label{eq:attpotential}
    U_{att} = \begin{dcases}
                \frac{1}{2}\zeta \norm{P - P_t}^2, &\text{if} \norm{P - P_t} \leq \rho, \\
                \zeta \rho \norm{P - P_t}, &\text{if} \norm{P - P_t} > \rho.
                \\
    \end{dcases}
\end{equation}
Where $\zeta$ and $\rho$ are the coefficients to describe the strength of potential field. Similar to the calculation of potential field in physics, the attractive force is given by taking the partial derivative of the attractive potential field:
\begin{equation} \label{eq:attforce}
    F_{att} = -\frac{\partial U_{att}}{\partial p}
    \\
    = \begin{dcases}
                -\zeta (P - P_t), &\text{if} \norm{P - P_t} \leq \rho, \\
                -\zeta \rho \frac{P - P_t}{\norm{P - P_t}},& \text{if} \norm{P - P_t} > \rho.
                \\
    \end{dcases}
\end{equation}
The resultant attractive force is constant when the robot is far away from the target. It decreases with decreasing distance to the target. Considering multiple obstacles surrounding the robot, let $P_{obs,i}$ be the $i^{th}$ obstacle position in the global frame, the repulsive potential can be calculated as 
\begin{equation} \label{eq:reppotential}
    U_{rep,i} = \begin{dcases}
                \frac{1}{2}\eta (\frac{1}{\norm{P - P_{obs,i}}} - \frac{1}{d_c})^2, &\text{if} \norm{P - P_{obs,i}} \leq d_c, \\
                0, &\text{if} \norm{P - P_{obs,i}} > d_c.
                \\
    \end{dcases}
\end{equation}
Where $\eta$ is the repulsive potential field strength coefficient, $d_c$ is the distance which the function is valid for. Similarly, repulsive force can be represented by
\begin{equation} \label{eq:repforce}
    F_{rep,i}
    \\
    = \begin{dcases}
                -\eta \left(\frac{1}{\norm{d}} - \frac{1}{d_c}\right)  \frac{d}{\norm{d}}, &\text{if} \norm{d} \leq d_c, \\
                0, &\text{if} \norm{d} > \rho,
                \\           
    \end{dcases}
\end{equation}
where $ d = P - P_{obs,i} $. In real practice, global knowledge of environment is usually unavailable prior to the commencement of mission. Therefore, the position of obstacle is usually derived from the sensors mounted on the robot. For example, $n$ laser rangefinders are mounted on a robot. The \(i^{th}\) rangefinder has an angle \(\theta_i\) with respect to the robot frame. For real-time processing, the obstacle can be estimated as 
\begin{equation}
    P_{obs,i} = P + d_i * \begin{bmatrix}cos(\theta_i)\\sin(\theta_i) \end{bmatrix},
\end{equation}
where $d_i$ is the reading of the $i^{th}$ sensor. Finally, the overall artificial potential force is obtained by
\begin{equation}
    F_{total} = F_{att} + \sum F_{rep,i}.
\end{equation}

\subsection{Wall-Following Method}
Wall-Following Method estimates the tangent and normal of the obstacle surface. The robot generates normal speed $v_{normal}$ with a simple PID controller to keep a certain distance with the obstacle. The moving speed $v_{tangent}$ is fixed with the tangent direction of obstacle surface. The velocity is hence given by
\begin{equation}
    v_{total} = v_{normal} + v_{tangent}.
\end{equation}

Another important factor in WFM is the decision on the wall-following direction. With historical trajectory look-back, same wall-following direction decision will be avoided when the robot revisits the position recorded in historical trajectory.

\subsection{Situation Awareness}
\begin{figure}[!h]
    \centering
    \includegraphics[width=0.80\linewidth]{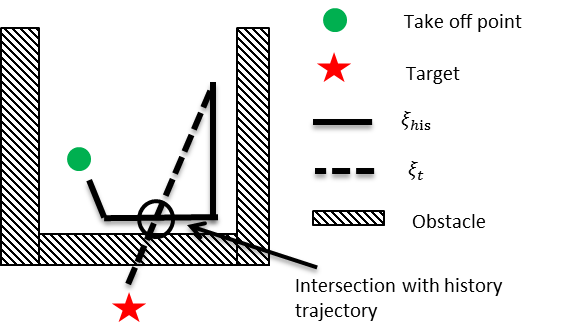}
    \caption{obstacle-less APF path intersects with history trajectory.}
    \label{fig:intersect_with_his_tra}
\end{figure}

With improvements from previous works described in the earlier sections, an enhanced situation awareness is proposed with capability to avoid the previously explained failure cases. We modified the existing conditions and added history look-back for both APF mode and WFM mode.\\ The system will switch to APF from WFM with the simplified equations shown as follow:
\begin{enumerate} [label=\roman*.]
    \item $\abs{\theta_{goal}-\theta_0}>\frac{\pi}{2} $,
    \item $\xi_t \cup \xi_{his} = \emptyset $,
\end{enumerate}
where $\xi_t$ is the robot-to-target straight line path and $\xi_{his}$ is the historical trajectory.
Both conditions have to be satisfied in order to trigger the WFM mode. When the robot is moving far away from the goal, the first condition tells that continuing Wall-Following Method may cause a longer trajectory. The set $\xi_t$ simply draws a straight line from the current position to the goal. The robot will switch to APF mode only when the intersection between this line and historical trajectory is empty. Otherwise, the robot has a greater chance to revisit the previous path so switching to another mode is not suggested. An illustration when robot fails to meet the second condition is shown in Fig. \ref{fig:intersect_with_his_tra}.  
\\Switching conditions from APF to WFM are given by:
\begin{enumerate} [label=\roman*.]
    \item $F_{total} < F_{th}$,
    \item If $p = p(t_0)$ and $v = v(t_0)$, $t_0<t_m$
\end{enumerate}
where $t_m$ records the time when the last local minimum appears, p is the current position, $p(t_0)$ and $v(t_0)$ are the position and moving direction at time $t=t_0$. \\The WFM mode will be triggered when either condition is satisfied. The first condition checks the local minimum. Whenever it happens, $t_m$ will be updated as the current time. The second condition is the key to avoid the revisit of historical path in APF mode. If in this mode, the robot finds that its current status coincides (both position and moving direction) with the history frame at time $t=t_0$, then continuing APF mode should not navigate itself to the local minimum recorded at $t=t_m$ again. In other words, the robot will switch to WFM if last local minimum with $t=t_m$ happens after current position with history record at time $t=t_0$ . If we do not trigger the mode in this case, the robot will likely make the same decision as it did from time $t_0$ to $t_m$. Hence the repetitive path will happen. 

\section{simulation result with complementary sensors}

\subsection{Sensing System}
In real life application, the information of the surrounding obstacles is often limited by the sensor constraints. In consideration of the implementation difficulty, cost and accuracy, we choose infrared sensor as sensing system. Infrared sensor is a time-of-flight(TOF) rangefinder with a small FOV. WLOG, the sensing system in the simulation contains 8 rangefinders equally distributed to get the 360-degree view of the surrounding environment. 

\subsection{Surface Estimation in MWF Mode}
\begin{figure}[!h]
\centering
    \includegraphics[width=0.80\linewidth]{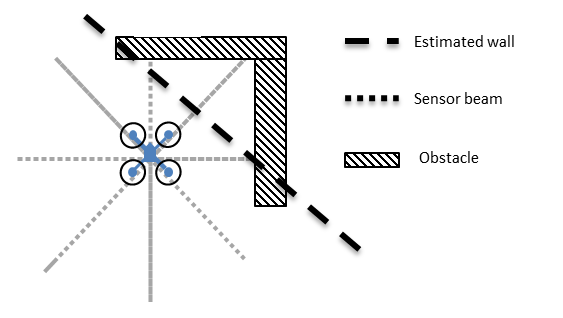}
    \caption{Surface estimation with rangefinder.}
    \label{fig:surface_estimation}
\end{figure}

In MWF mode, moving direction is determined by tangent of wall surface. The tangent of wall surface is estimated by selecting the tangent of nearest wall detected by the sensor. Each wall will be estimated by the following condition:
\begin{equation} \label{eq:wallestimation}
    d_{ij} = \frac{\frac{1}{2}d_id_jsin(\theta_{ij})}{\sqrt{d_i^2+d_j^2-2d_id_jcos(\theta_{ij})}}, \text{if }\theta_{ij}<\frac{\pi}{2},
\end{equation}
where $d_i$, $d_j$ and $\theta_{ij}$ are the readings of $i^{th}$, $j^{th}$ sensor and mounting angle between them. The minimum of $d_{ij}$ indicates the wall closest to robot and hence the tangent of this wall is chosen as the moving direction. Fig. \ref{fig:surface_estimation} gives an illustration of wall estimation.
\\
\subsection{Simulation Result}
\begin{figure}[!h]
\centering
    \includegraphics[width=0.90\linewidth]{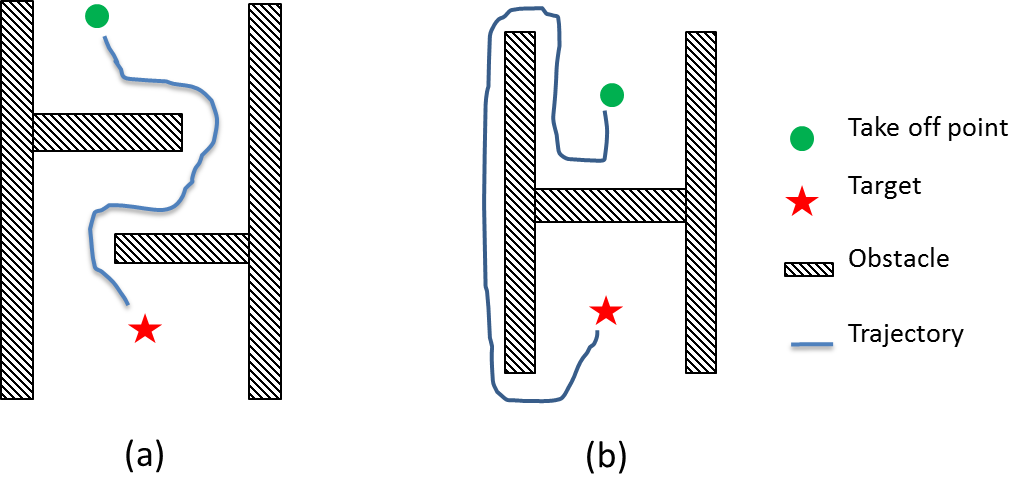}
    \caption{(a) Result of path planning when there is a straight path (b) Result of H shape obstacle}
    \label{fig:simulation1}
\end{figure}

The simulation is performed in RVIZ, a simulation tool based on Robotics Operating System (ROS). To validate the effectiveness of MWF-APF method, Fig. \ref{fig:simulation1} gives two different cases. In Fig. \ref{fig:simulation1}(a), the obstacles leave small openings in the corridor so that the robot can take a simple detour through the corridor. The planned path is straightforward as the heading of the robot does not change too much. In Fig. \ref{fig:simulation1}(b), it is more complex with the corridor closed. The robot has to make a u-turn and move outside corridor in order to navigate to the target. The robot is capable of planning different path based on the situation.

\begin{figure}[!h]
\centering
    \includegraphics[width=0.80\linewidth]{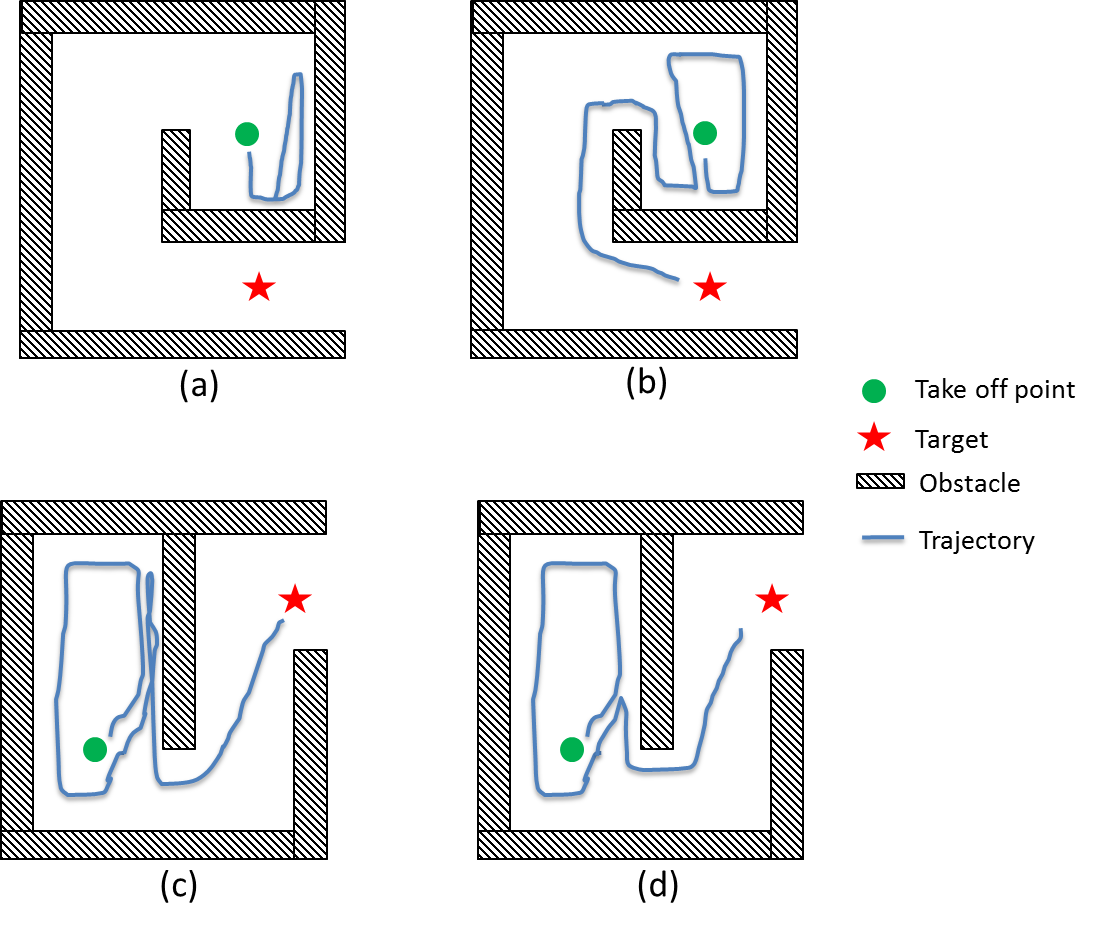}
    \caption{(a)Result of traditional WFM-APF method. (b)Result of MWF-APF method. (c)Result of WFM-APF method with only historical look-back in WFM mode (d)Result of MWF-APF method}
    \label{fig:simulation2}
\end{figure}

A comprehensive comparison with the existing methods is shown in Fig. \ref{fig:simulation2}. In Fig. \ref{fig:simulation2}(a) and Fig. \ref{fig:simulation2}(b), we compare the traditional memory-less WFM-APF method with our MWF-APF method. The result shows that, with the new method the robot can navigate to the goal point while with the traditional method it fails. Fig. \ref{fig:simulation2}(c) shows the case where historical trajectory is only considered in WFM mode. Our method shown in Fig. \ref{fig:simulation2}(d) makes a shorter path than the method in Fig. \ref{fig:simulation2}(c).

\section{Experiment Result}
In this section the proposed MWF-APF method is implemented on a quad-rotor. A short video of the experiment can be viewed at \url{https://youtu.be/E1h35GH9-tU}.
A series of experiments are conducted in an indoor environment where motion capture system is used to provide position information of the UAV. The test field of the size 5m * 5m is set up to form 3 obstacles in the center. 

\begin{figure}[!h]
\centering
    \includegraphics[width=0.80\linewidth]{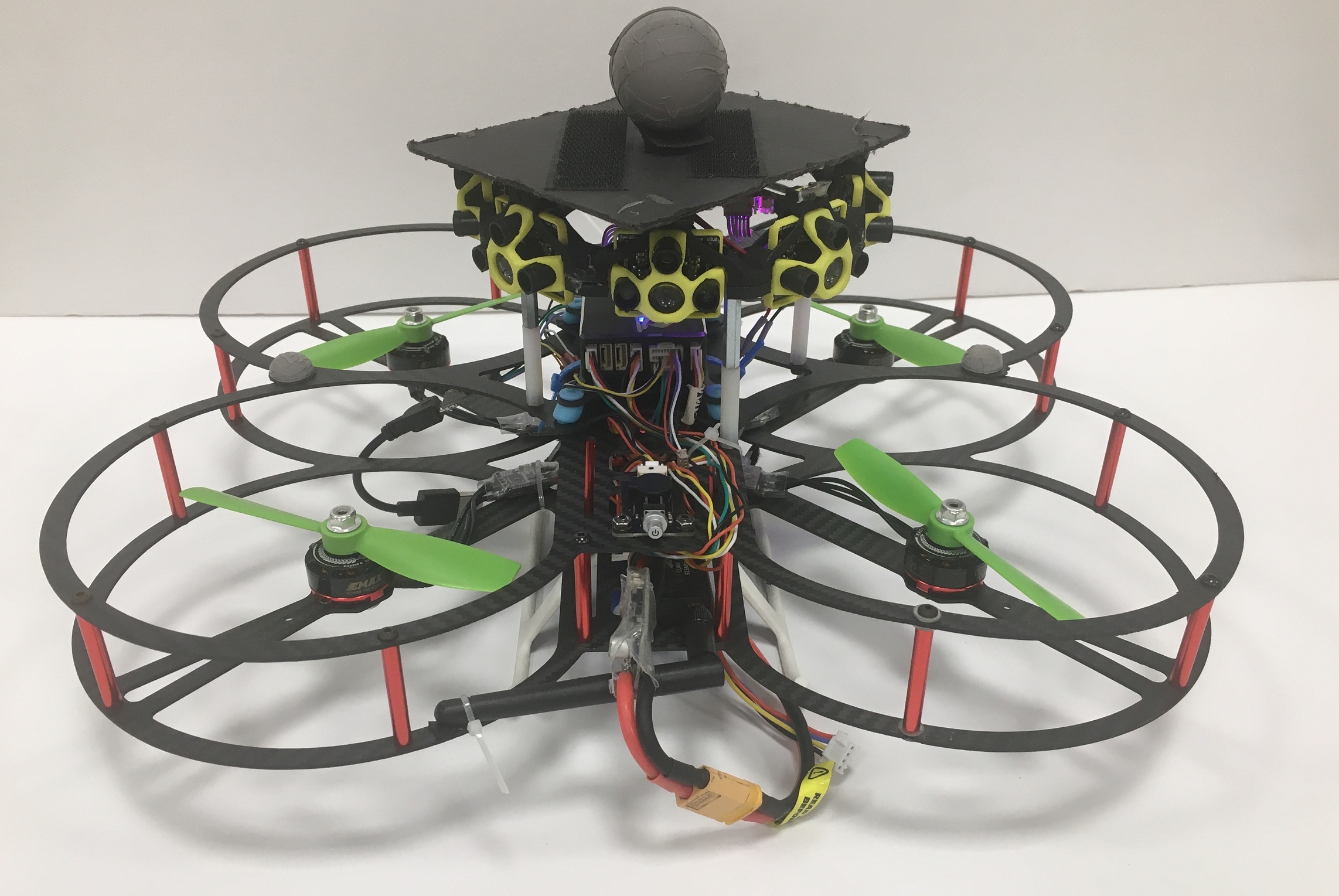}
    \caption{UAV Mechanical Overview.}
    \label{fig:uav}
\end{figure}

A quad-rotor is constructed that integrates distance sensors, a flight controller and an on-board computing unit. A photo of the quad-rotor is shown in Fig. \ref{fig:uav}. A group of 5 infrared time-of-flight (TOF) distance sensors is mounted above the UAV main frame for unobstructed distance measurement to 5 directions. The sensors are arranged such that they face $-90^{o}$, $-45^{o}$, $0^{o}$, $45^{o}$ and $90^{o}$ relative to vehicle front, so that a wide range of view is achieved.

\begin{figure}[!h]
\centering
    \includegraphics[width=0.9\linewidth]{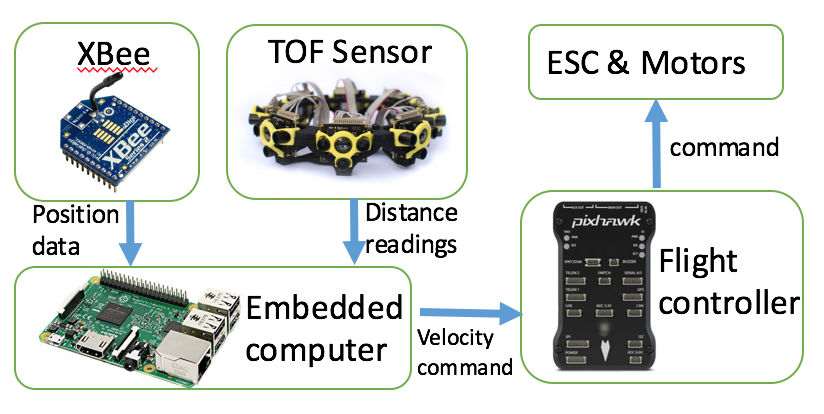}
    \caption{UAV System Overview.}
    \label{fig:system}
\end{figure}

The system overview of UAV is depicted in Fig. \ref{fig:system}. The flight controller runs on-board PX4 firmware for position and attitude control. Position data is sent from an off-board motion capture system to the UAV through wireless XBee receiver. The distance sensor readings are sent to a 900MHz quad-core ARM embedded computer which runs Ubuntu and ROS. The proposed obstacle avoidance algorithm is executed in real time in ROS nodes which takes the UAV real-time position and distance sensor readings as inputs and outputs UAV's target velocity in lateral axes.

\begin{figure}[!h]
\centering
    \includegraphics[width=0.225\textwidth]{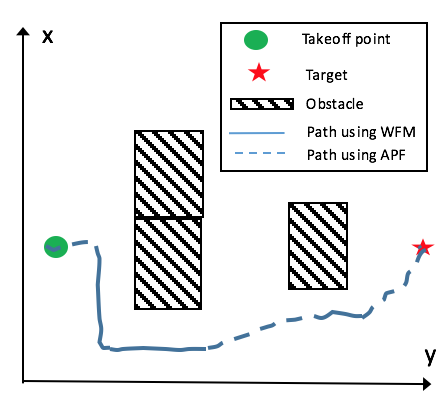}
    \includegraphics[width=0.225\textwidth]{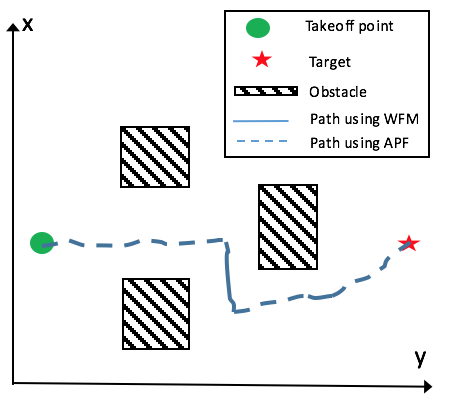}
    \caption{Indoor experiment plot showing the UAV path in X/Y axes: UAV is able to switch between APF and WFM and reach target with an efficient path.}
    \label{fig:exp-path}
\end{figure}

Multiple flight tests are conducted. During each flight, the UAV has to fly to a target position which is obstructed by multiple boxes. Obstacle profile is unknown prior to the flight and is detected online by the distance sensors. The UAV's lateral velocity is commanded by the MWF-APF algorithm while its altitude is maintained constant at 1m by the altitude controller. In each flight test, the UAV is able to reach the target position without colliding into the obstacles. Plots of some flight tests are shown in Figure.\ref{fig:exp-path}. It is observed that the proposed method is able to switch between APF and WFM algorithms intelligently, allowing the UAV to effectively reach the target. 
 
\section{Conclusion} \label{sec:conclusion}

This paper presented a feasible and computationally efficient real-time path planning method for UAV with complementary sensors. The algorithm, namely memory based wall following-artificial potential field (MWF-APF) method, combines Wall-Following Method and Artificial Potential Field method with enhanced situation awareness. A comprehensive review of the previous path planning methods was given. To overcome the limitations of the existing methods, the new method includes the historical trajectory look-back on both APF and WFM modes. It provides an alternative solution to the endless loop problem of WFM and the local minimum problem of APF. The revised situation awareness is able to avoid repetitive path with guaranteed success of navigation compared to existing methods on WFM and APF. To the best of author's knowledge, the proposed situation awareness has better performance than existing works on method combining WFM and APF in terms of generated path length and success of navigation.
\\
At the end of this paper, different simulation scenarios were given to show the robustness of the WFM-APF method in robot path planning. A collision-free UAV experiment was provided to validate its capability to complete real life mission. It was shown that only a few low-cost rangefinders are needed to execute the MFM-APF method and to complete the navigation task.

\section*{Acknowledgement}
The author would like to thank Mr. Nguyen Pham Nhat Thien Minh for the great many suggestions during the planning and development of this research work.
\balance
\bibliographystyle{IEEEtran}
\bibliography{IEEEabrv,./references}

\end{document}